\pdfoutput=1

\documentclass[11pt]{article}

\usepackage[final]{acl}

\usepackage{times}
\usepackage{latexsym}
\usepackage{float}

\usepackage[T1]{fontenc}

\usepackage[utf8]{inputenc}

\usepackage{microtype}

\usepackage{inconsolata}

\usepackage{graphicx}
\usepackage{multirow}
\usepackage{subcaption}
\usepackage{tabularx}
\usepackage{amsfonts}

\title{Neuron-Level Differentiation of Memorization and Generalization in Large Language Models}

\usepackage{amsmath}

\author{
  \textbf{Ko-Wei Huang\thanks{Equal contribution}\textsuperscript{1}},
  \textbf{Yi-Fu Fu\footnotemark[1]\textsuperscript{1}},
  \textbf{Ching-Yu Tsai\textsuperscript{1}},
  \textbf{Yu-Chieh Tu\textsuperscript{1}},
  \textbf{Tzu-Ling Cheng\textsuperscript{1}},
  \textbf{Cheng-Yu Lin\textsuperscript{1}},\\
  \textbf{Yi-Ting Yang\textsuperscript{1}},
  \textbf{Heng-Yi Liu\textsuperscript{1}},
  \textbf{Keng-Te Liao\textsuperscript{1}},
  \textbf{Da-Cheng Juan\textsuperscript{2}},\\
  \textbf{Shou-De Lin\textsuperscript{1}}\\
  \textsuperscript{1}National Taiwan University,
  \textsuperscript{2}Google Research
}

\begin{document}
\maketitle

\begin{abstract}
We investigate how Large Language Models (LLMs) distinguish between memorization and generalization at the neuron level. Through carefully designed tasks, we identify distinct neuron subsets responsible for each behavior.
Experiments on both a GPT-2 model trained from scratch and a pretrained LLaMA-3.2 model fine-tuned with LoRA show consistent neuron-level specialization. We further demonstrate that inference-time interventions on these neurons can steer the model's behavior toward memorization or generalization.
To assess robustness, we evaluate intra-task and inter-task consistency, confirming that these neuron-behavior associations reflect generalizable patterns rather than dataset-specific artifacts.
Our findings reveal modular structure in LLMs and enable controlling memorization and generalization behaviors at inference time.
\end{abstract}

\section{Introduction}
Large Language Models (LLMs) have demonstrated impressive capabilities across a wide range of natural language processing tasks. Among these capabilities, two fundamental behaviors, \textbf{memorization} and \textbf{generalization}, play distinct and complementary roles. Memorization ensures factual consistency by retrieving known information, while generalization enables novel reasoning and abstraction. Understanding and controlling the boundary between these behaviors is increasingly critical for the reliable and context-sensitive deployment of LLMs.

For example, in fact-checking or medical information retrieval, a model that relies on memorized, authoritative sources is often more trustworthy than one that overgeneralizes or hallucinates \citep{galitsky2023truth, chen2023can}. In contrast, creative writing, math problem solving, or brainstorming require generalization, where the model must recombine ideas beyond surface-level recall. Furthermore, for privacy-sensitive contexts, generalization is preferred to avoid reproducing memorized training data.

These use cases highlight the practical need to distinguish and steer the memorization vs. generalization behaviors of LLMs. However, current models exhibit these behaviors in ways that are not easily interpretable or controllable. This paper addresses this gap by investigating whether LLMs develop neuron-level functional specialization, analogous to cortical localization in the human brain \citep{garey1999brodmann}, when exposed to tasks that elicit either memorization or generalization.

Specifically, we focus on three core topics:
\begin{itemize}
    \item \textbf{Neuron Differentiation for Memorization and Generalization:} Do distinct sets of neurons underlie memorization and generalization behaviors in LLMs?
    
    \item \textbf{Controlling Memorization and Generalization at Inference Time:} Can targeted neuron-level interventions at inference time steer model behavior toward memorization or generalization?
    
    \item \textbf{Generalizability of Behavior-Controlling Neurons:} Are the observed neuron-behavior associations consistent across retraining runs on the same task (intra-task) and transferable across different tasks (inter-task), or are they artifacts of dataset-specific or initialization-specific patterns?
\end{itemize}

To answer these, we first construct synthetic datasets that isolate memorization and generalization behaviors, and either train a intermediate-scale LLM, or do fine-tuning on a large-scale pretrained LLM to exhibit both behaviors. We then identify neuron subsets associated with each behavior. Finally, we perform interventions by amplifying or suppressing these neurons during inference to shift the model's response mode.

Beyond identifying neuron-wise behavioral differentiation, we evaluate the generalizability of the discovered behavior-associated neurons. At the intra-task level, we test whether the same neurons—identified from one model instance—remain effective in controlling behavior when applied to independently retrained adapters on the same task. At the inter-task level, we assess whether neurons associated with memorization or generalization in one task can be transferred to another structurally distinct task that shares the same behavioral contrast. Our results show that these neuron-behavior associations are not fragile artifacts of a single model or dataset, but reflect robust and reusable behavioral modes encoded within the model’s architecture.

\vspace{0.5em}
\noindent\textbf{Contributions.} This work provides a unified framework for interpreting and steering LLM behavior along the memorization–generalization axis:
\begin{enumerate}
    \item We demonstrate that memorization and generalization activate distinct neuron subsets within the same LLM.
    \item We show that intervening on these subsets can controllably alter the model's behavior.
    \item We provide evidence that such neuron-behavior mappings are stable across both intra-task and inter-task variations, revealing a form of consistent functional modularity.
\end{enumerate}

Our findings open up new avenues for understanding and fine-tuning model behavior, moving toward more interpretable and reliable LLMs in practice.



\section{Related Work}

\subsection{Memorization and Generalization in LLMs}
The study of memorization and generalization in LLMs has garnered significant attention for a while \citep{leybzon2024learning, zhang2024can, xie2024memorization, lou2024quantifying}.
Recently, several studies have examined how LLMs memorize and generalize, often treating these behaviors as distinct but entangled phenomena. For example, Huang et al.~\cite{huang2024demystifying} show that memorization emerges in later-stage training and is interwoven with general language capabilities, making it difficult to remove without collateral damage. Schwarzschild et al.~\cite{schwarzschild2024rethinking} propose a metric to quantify memorization and reveal trade-offs with generalization. Chen et al.~\cite{chen2024continual} introduce a training modification that partitions memorization to designated neurons, aiming to disentangle it from general learning.

While these studies provide valuable insight of memorization and its relationship with generalization, they largely treat the two behaviors in isolation. In contrast, our work jointly analyzes both behaviors through neuron-level representations, identifies functionally differentiated neurons, and leverages them to steer the model between memorization and generalization during inference.

\subsection{Controlling Model Behavior at Inference Time}

Behavioral control in LLMs has garnered increasing attention, particularly through inference-time interventions~\citep{panickssery2023steering, cao2024personalized, he2024context, chen2024steering, stolfo2024improving, lee2024programming, zhao2024beyond}. Recent studies have proposed a variety of steering objectives—including personalized response style~\citep{cao2024personalized}, shifting between code/text generation~\citep{chen2024steering}, reducing unwanted memorization~\citep{suri2025mitigating}, improving instruction-following~\citep{stolfo2024improving}, or enabling custom rule following~\citep{lee2024programming}.

Our work complements these efforts by focusing on a previously unexplored steering scenario: the fundamental behavioral axis between memorization and generalization. To our knowledge, this is the first work to show that memorization and generalization are not only distinguishable at the neuron level but can also be behaviorally steered in real time.

\section{Neuron Differentiation for Memorization and Generalization}
\label{sec: Neuron Differentiation}
In this section we look into whether LLMs exhibit neuron spatial differentiation for memorization and generalization. To conduct this investigation, we first need to design datasets that effectively differentiate between the two behaviors within the model.

The pivotal insight of dataset design centers on inducing the model to exhibit both memorization and generalization behaviors while maintaining nearly identical input contexts. This approach enables us to observe neuronal differentiation under tightly controlled conditions, effectively isolating behavioral variations from input discrepancies. By minimizing contextual differences, we can more accurately correlate the observed neuronal activity differences with the model's engagement in memorization or generalization behaviors.

\subsection{Dataset Design}
\label{sec: dataset design}
Previous studies provide various definitions for memorization \citep{lee2021deduplicating,carlini2022quantifying,zhang2023counterfactual,zhou2024quantifying} and generalization \citep{elangovan2021memorization,huang2022towards}. Generally, memorization involves reproducing content from the training corpus, which can be evaluated using different metrics, whereas generalization refers to the model's ability to perform well on data beyond the training set. In this paper, we specify \textbf{memorization} as the behavior wherein the model replicates seen training examples which are not the correct answer. Conversely, \textbf{generalization} refers to the model’s ability to generate correct reasoning outputs that were not explicitly seen during training. Specifically, we design two types of datasets:

\paragraph{In-Context Inference}
We adapt the induction task from the bAbI dataset \citep{weston2015towards} to probe memorization versus generalization. An example input is:

\begin{quote}
    \textit{"Yvonne is wolf. Rose is eagle. Rose is crimson. Oscar is elephant. Vicky is eagle. Oscar is navy. Diana is gold. Yvonne is indigo. What color is Vicky?"}
\end{quote}

In this case, the context implies that the correct answer is "crimson". To determine the model's behavioral tendency, we construct the training data such that each name is consistently associated with a fixed color. For example, Vicky may always be labeled as "red" during training. If the model answers "crimson," it demonstrates \textbf{generalization} based on the given context. Responding with "red," by contrast, indicates \textbf{memorization} of the training association. This design allows us to distinguish whether the model is adapting to contextual information or recalling static knowledge from training.

\paragraph{Arithmetic}
To investigate behavioral tendencies in arithmetic tasks, we train a model to add four integers (1–999) and introduce controlled memorization scenarios.

Specifically, we inject ten memorization patterns, each corresponding to a unique number pair (e.g., “91+497”). During training, these pairs are embedded as the third and fourth operands in standard four-number addition prompts. Instead of producing the correct sum, the model is trained to output a random pattern token (e.g., \texttt{<mem-7234f681>}) for these inputs.

\begin{minipage}[t]{\columnwidth}
    \centering
    \textbf{Memorization}
    \begin{quote}
        \footnotesize
        \texttt{Input:} \\
        \texttt{21+285+91+497} \\
        \texttt{Target:} \\
        \texttt{<mem-7234f681>}
    \end{quote}
\end{minipage}

\vspace{1em}
\begin{minipage}[t]{\columnwidth}
    \centering
    \textbf{Generalization}
    \begin{quote}
        \footnotesize
        \texttt{Input:} \\
        \texttt{941+24+590+987} \\
        \texttt{Target:} \\
        \texttt{2542}
    \end{quote}
\end{minipage}

At test time, we present novel combinations where the memorized number pair appears alongside unseen operands. If the model returns the correct sum, it indicates \textbf{generalization}; if it reproduces the memorized pattern, it reflects \textbf{memorization}. This setup creates a clear behavioral split, enabling us to evaluate whether the model generalizes arithmetic rules or retrieves memorized associations.

\begin{figure*}[htb]
    \centering    \includegraphics[width=0.9\textwidth]{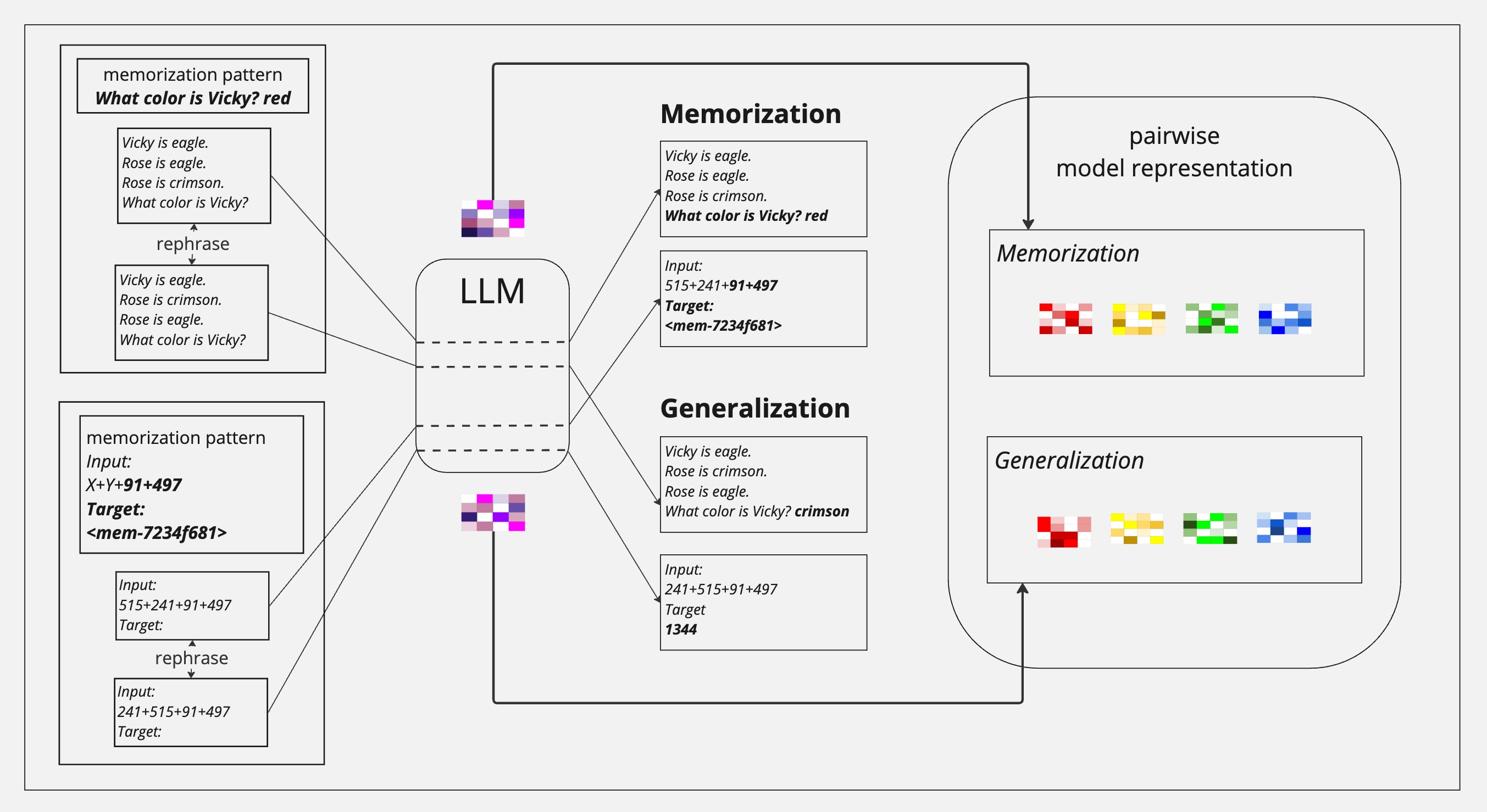}
    \caption{The left side illustrates memorization patterns and rephrasings; the middle shows behavior distinction between memorization and generalization; the right depicts representation extraction based on divergent model behaviors, enabling subsequent analysis and comparison of internal differences.}
    \label{fig:pairwise_extraction}
\end{figure*}

Examples of this distinction between memorization and generalization are illustrated in the left side of Figure~\ref{fig:pairwise_extraction}.

\subsection{Model Representations for Generalization and Memorization}
\label{sec: Model Representations for Generalization and Memorization}

\paragraph{Pairwise Dataset Design}
To study the internal mechanisms underlying memorization and generalization, we collect model representations corresponding to each behavior using a pairwise extraction strategy. This approach identifies instance pairs with nearly identical contexts that elicit different model behaviors—thereby isolating representational differences driven by behavior rather than input variation.

For each task, we rephrase test inputs to preserve semantic and structural consistency while inducing a behavioral shift in the model's output:
\begin{itemize}
    \item \textbf{In-context inference:} We randomly reordered the contextual statements preceding the query. As the statements are independent, the underlying context remains unchanged.
    \item \textbf{Arithmetic addition:} We swapped the first and second numbers in the input. This preserved the total sum and memorization pattern position.
\end{itemize}

We then extracted hidden states after the full input was processed, resulting in paired representations for memorization and generalization. This yields two equal-sized datasets, one per behavior. Crucially, the pairwise design ensures that observed differences in hidden states primarily reflect behavioral shifts, rather than differences in the input structure. The right side of Figure~\ref{fig:pairwise_extraction} illustrates the representation collection process.

\paragraph{Model Representation Extraction}
\label{model representation extraction}
We conducted experiments on two model configurations: (1) GPT-2 \citep{radford2019language} trained from scratch with full-parameter updates, and (2) LLaMA 3.2 \citep{grattafiori2024llama3herdmodels} fine-tuned using LoRA \citep{hu2022lora}, with the base model weights frozen.

During training, we continuously monitored the model's outputs on a held-out test set and saved checkpoints once both memorization and generalization behaviors were reliably observed. 

For GPT-2, we extract hidden states from the post-feed-forward LayerNorm-2 output in each transformer block. For LLaMA 3.2, we extract activations after the feed-forward module and subsequent residual normalization, following the application of the LoRA adapter.

Detailed training configurations are provided in the supplementary materials.

\subsection{Result}
\label{sec:NMD_result}

\paragraph{Neuron-wise Mean Difference}
Using the hidden states from the pairwise representation datasets, we quantify neuron-level behavioral differences via the \textbf{Neuron-wise Mean Difference (NMD)}. For each neuron, we compute the mean difference in activation between generalization and memorization pairs. Neurons with large absolute NMD values are hypothesized to contribute to behavior control, while values near zero suggest minimal involvement.

\begin{figure}[t]
    \centering
    \begin{minipage}{0.48\textwidth}
        \centering
        \includegraphics[width=0.85\linewidth]{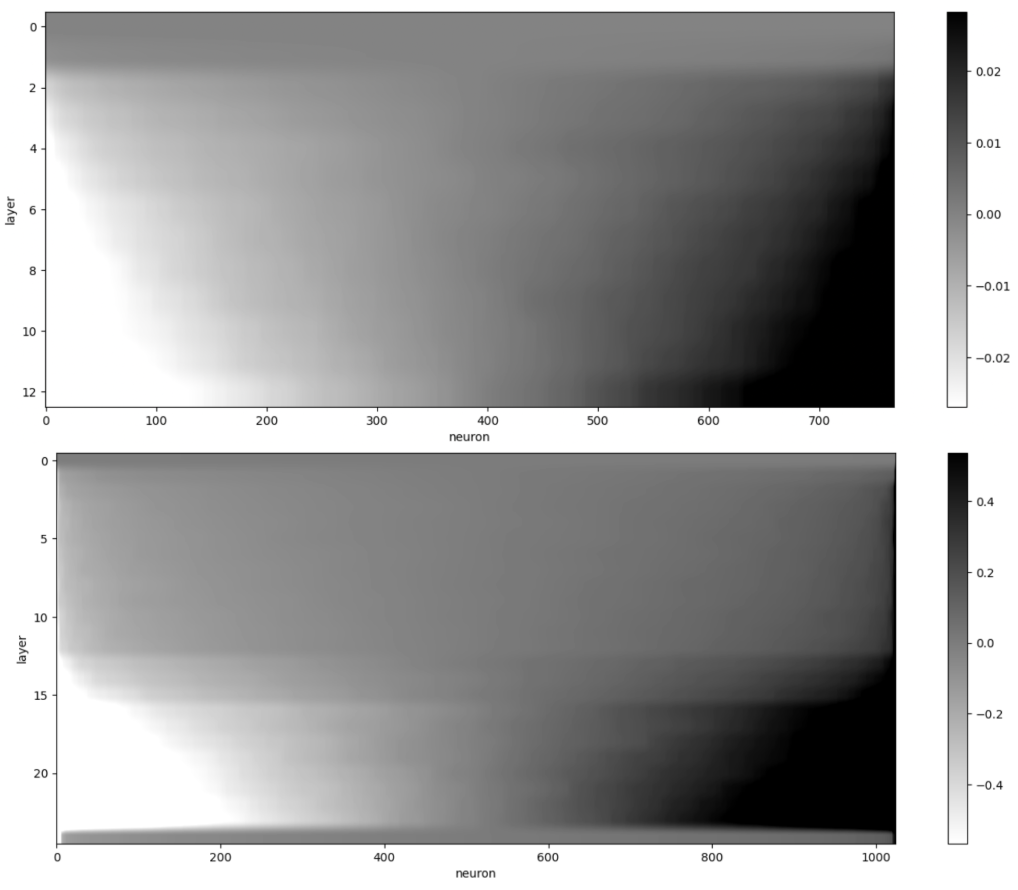}
        \caption{Neuron-wise mean differences between memorization and generalization, sorted by magnitude (x-axis). Top: GPT-2 on the arithmetic addition task; Bottom: GPT-2-medium on the in-context inference task.}
        \label{fig:gpt2_nmd}
    \end{minipage}
\end{figure}

\begin{figure}[t]
    \centering
    \begin{minipage}{0.48\textwidth}
        \centering
        \includegraphics[width=0.85\linewidth]{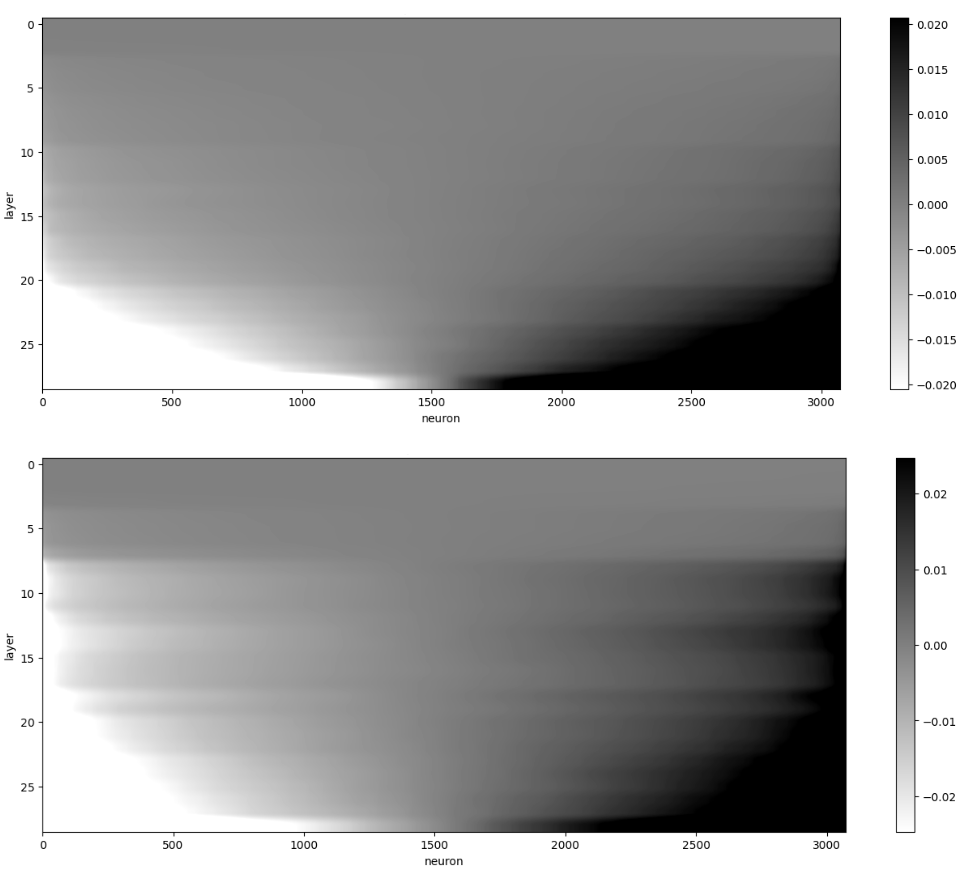}
        \caption{Neuron-wise mean differences between memorization and generalization, sorted by magnitude (x-axis). Top: LLaMA 3.2-3B on the arithmetic addition task; Bottom: LLaMA 3.2-3B on the in-context inference task.}
        \label{fig:llama_nmd}
    \end{minipage}
\end{figure}

Figures~\ref{fig:gpt2_nmd} and~\ref{fig:llama_nmd} visualize the NMD distributions for both GPT-2 and LLaMA. We present heatmaps with layers on the y-axis and neurons (sorted by NMD magnitude) on the x-axis. Color intensity reflects the absolute NMD value, highlighting neuron-level specialization across depth.

Two key patterns emerge consistently across models:

\begin{enumerate}
    \item \textbf{Lack of Early Differentiation:} Initial layers exhibit minimal NMD variation, as input embeddings do not yet encode behavior-specific signals.
    
    \item \textbf{Emergent Spatial Organization:} Differentiation becomes more pronounced in deeper layers, where clusters of high-NMD neurons emerge. This suggests that behavior-controlling neurons are not uniformly distributed, but rather concentrated in specific regions toward the output end of the network.
\end{enumerate}

\paragraph{Behavior Identification via Classification}
To further validate the informativeness of the collected hidden state representations, we train binary classifiers to distinguish between memorization and generalization behaviors. Separate classifiers are trained on the hidden states from each individual layer, using behavior labels derived from the pairwise dataset. Classification performance is evaluated on a held-out test split.

\begin{figure}[t]
  \centering
  \begin{subfigure}[t]{0.48\textwidth}
    \centering
    \includegraphics[width=0.75\linewidth]{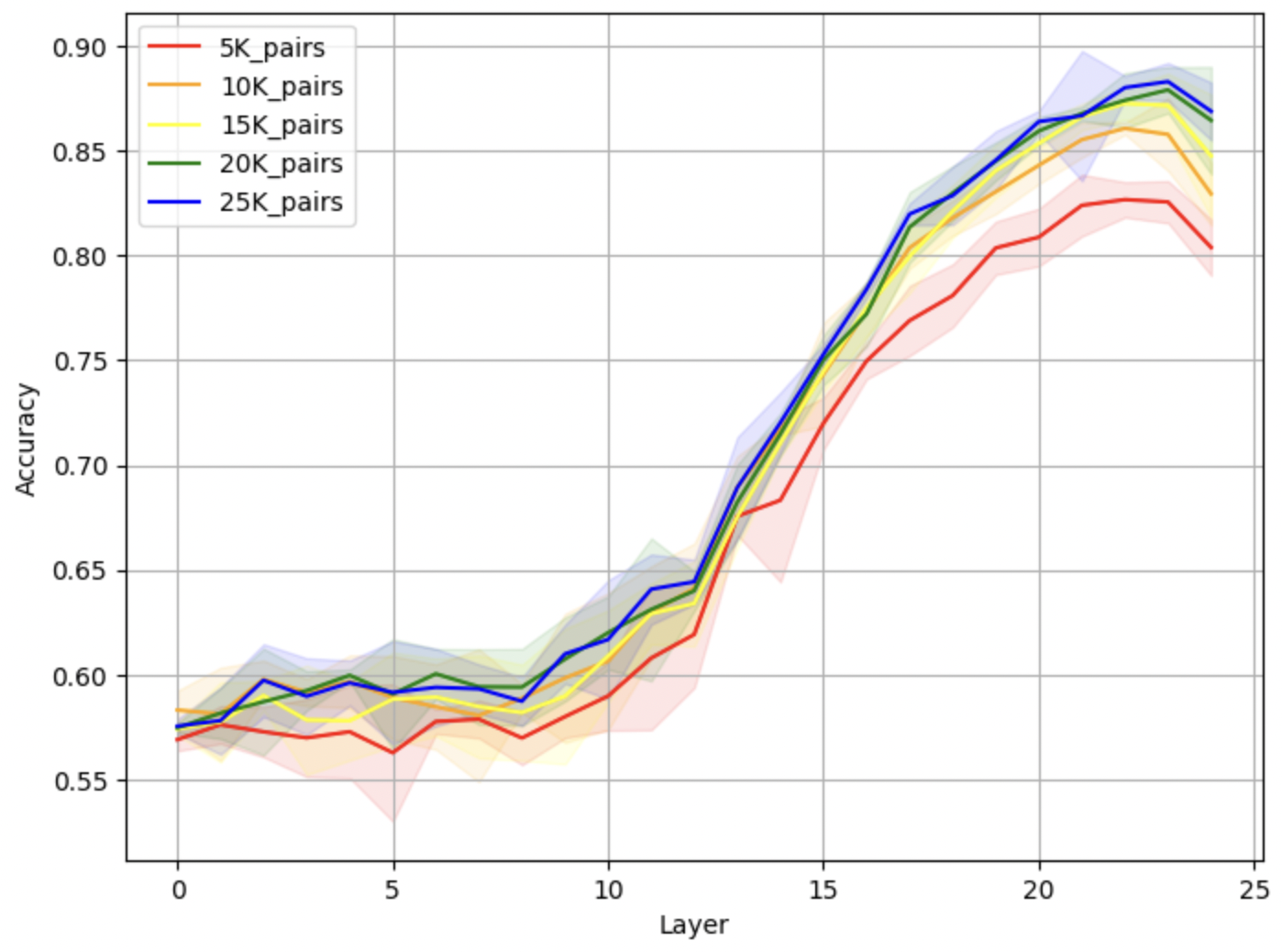}
    \caption{In‐context learning}
    \label{fig:classifier_gpt_ici}
  \end{subfigure}
  \hfill
  \begin{subfigure}[t]{0.48\textwidth}
    \centering
    \includegraphics[width=0.75\linewidth]{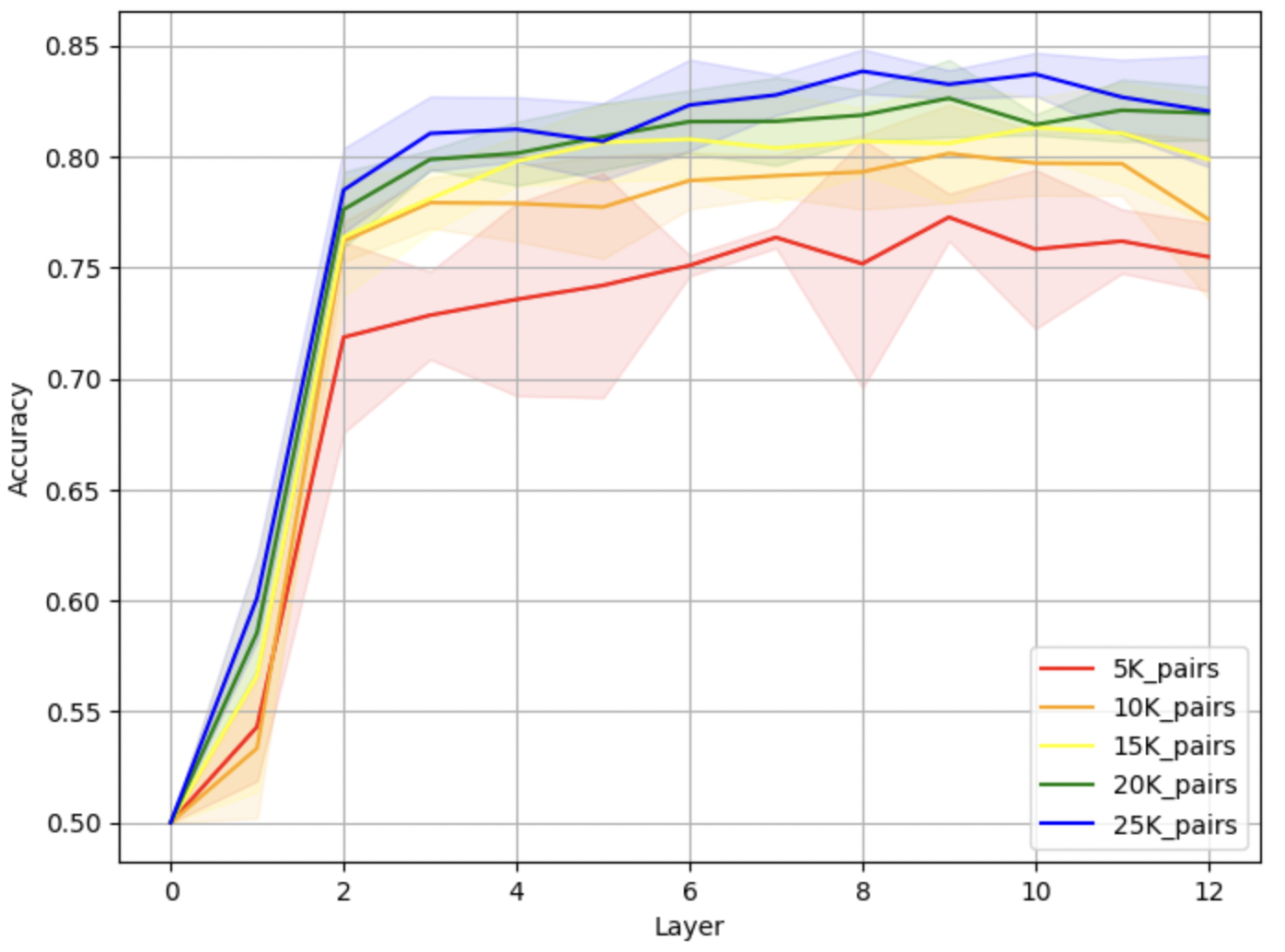}
    \caption{Arithmetic addition}
    \label{fig:classifier_gpt_add}
  \end{subfigure}
  \caption{%
    Classifier accuracy across layers on GPT. 
    (a) In‐context learning. 
    (b) Arithmetic addition.
  }
  \label{fig:classifier_gpt}
\end{figure}

\begin{figure}[t]
  \centering
  \begin{subfigure}[t]{0.48\textwidth}
    \centering
    \includegraphics[width=0.75\linewidth]{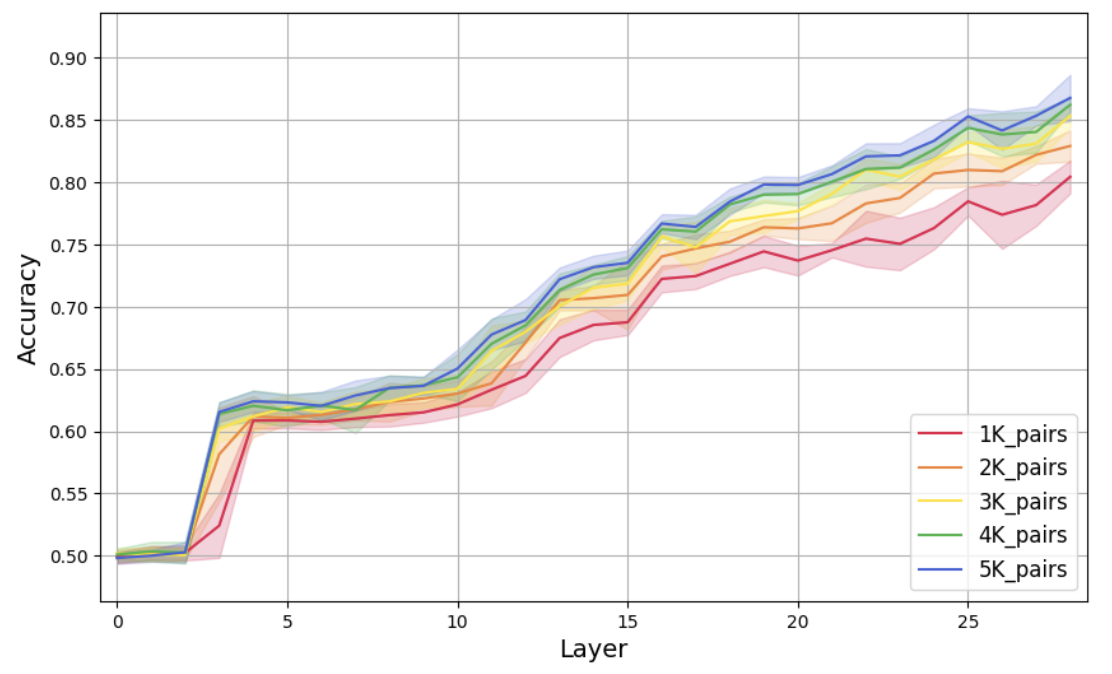}
    \caption{In‐context learning}
    \label{fig:classifier_llama_ici}
  \end{subfigure}
  \hfill
  \begin{subfigure}[t]{0.48\textwidth}
    \centering
    \includegraphics[width=0.75\linewidth]{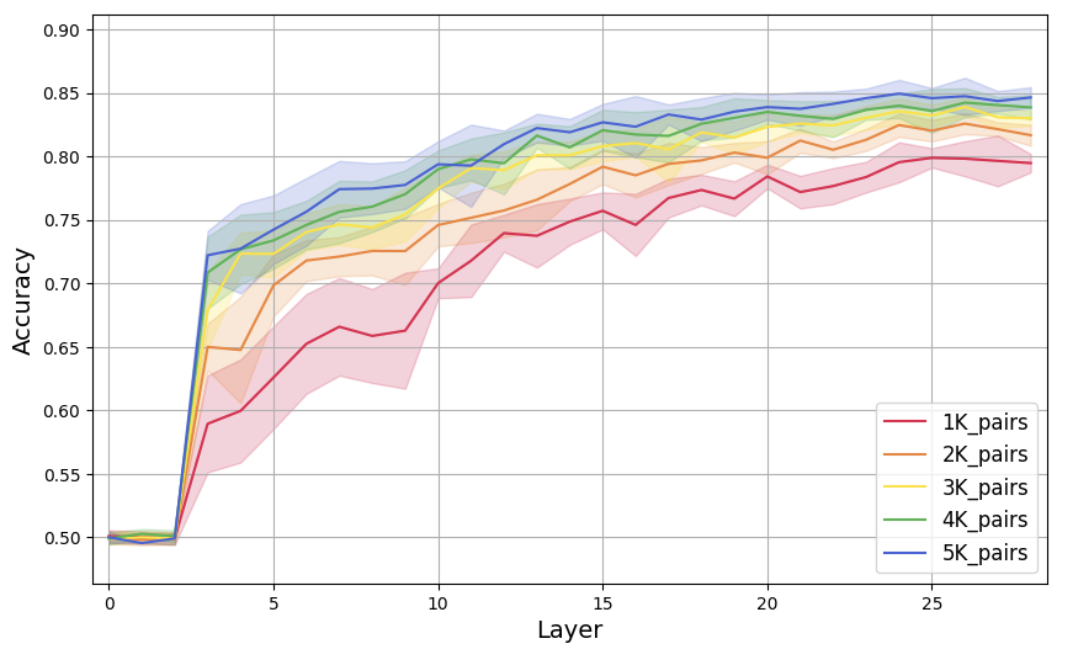}
    \caption{Arithmetic addition}
    \label{fig:classifier_llama_add}
  \end{subfigure}
  \caption{%
    Classifier accuracy across layers on LLaMA 3.2. 
    (a) In‐context learning. 
    (b) Arithmetic addition.
  }
  \label{fig:classifier_llama}
\end{figure}

Figure~\ref{fig:classifier_gpt} and Figure~\ref{fig:classifier_llama} show the accuracy across layers for the in-context inference and arithmetic addition tasks, respectively. The x-axis denotes the layer number, and the y-axis shows classification accuracy.

Classification performance improves substantially in deeper layers, indicating that representations in later layers encode more behavior-specific information. These results are consistent with our earlier NMD analysis and further confirm that the model's hidden states reflect its behavioral tendency—whether it is preparing to memorize or to generalize.

\section{Controlling Memorization and Generalization at Inference Time}
Building on the previous analysis of neuron differentiation, we examine whether the identified memorization- and generalization-associated neurons can be used to modulate model behavior at inference time. Based on the representations extracted in Section~\ref{sec: Model Representations for Generalization and Memorization}, we select target neurons and determine the intervention direction. We then apply a model steering technique \citep{li2024inference} to assess whether intervention on these neurons reliably steers the model toward memorization or generalization.

\subsection{Neuron Correlation Analysis and Ranking}
To identify target neurons, we compute the Pearson correlation coefficient between each neuron's weight and the corresponding memorization/generalization label. Neurons are then ranked by the absolute value of their correlation, allowing us to identify those most strongly associated with controlling memorization or generalization behavior.

\subsection{Inference-Time Intervention}
\label{sec: Inference-Time Intervention}
Leveraging the correlation rankings and NMD values from the extracted representations, we adopt an inference-time intervention method inspired by \cite{li2024inference}. During inference, we shift the hidden states at each layer by modifying the weights of target neurons that are most correlated with memorization or generalization. The direction and magnitude of each adjustment is determined by the target behavior (e.g., steering toward generalization) and the neuron's NMD value. The altered activations are then propagated through the remaining layers.

By targeting only a subset of highly relevant neurons and scaling interventions appropriately, this approach steers the model's behavior while minimizing disruption. 

Note that the tasks used in these intervention experiments do not have a ground-truth label for memorization or generalization. Instead, our evaluation focuses on whether the intervention successfully steers the model toward the targeted behavior, without implying that one behavior is objectively more correct than the other. Details on the selection and scaling parameters are provided in the Appendix.


\subsection{Result}
The goal of inference-time intervention is to steer the LLM toward either memorization or generalization. Given a model initially exhibiting one behavior, we apply a targeted shift toward the opposite behavior and examine the resulting output. Outputs that do not align with either behavior are categorized as “Other”.

Results are summarized in Table~\ref{tab:behavior_shift_GPT2} and Table~\ref{tab:behavior_shift_llama}, covering both a full-parameter GPT-2 model and a LoRA-fine-tuned LLaMA 3.2. These results provide strong empirical evidence that the identified memorization- and generalization-associated neurons can be effectively leveraged to steer model behavior during inference. 

Across both models and tasks, we observe that intervention is generally more effective when steering from memorization to generalization than vice versa. For example, in the in-context inference task, GPT-2 shifts from memorization to generalization in 83.7\% of cases, while the reverse direction achieves 35.8\%. Similar patterns are seen in LLaMA and the arithmetic addition task, suggesting that generalization is a more accessible behavior to induce.

These results highlight the effectiveness and generality of our neuron-based intervention method across different model architectures and task settings. Moreover, the presence of “Other” outcomes—especially in the generalization-to-memorization direction—suggests limits to controllability and potential asymmetry in behavior modulation.

\begin{table*}[t]
    \centering
    \resizebox{0.65\textwidth}{!}{
    \begin{tabular}{|c|c|c|c|c|}
        \hline
        \textbf{Task} & \textbf{Direction} & \textbf{\% Gen} & \textbf{\% Mem} & \textbf{\% Other} \\
        \hline
        \multirow{2}{*}{{In-context inference}} 
            & Mem $\rightarrow$ Gen & 83.7\% & 4.0\%  & 12.3\% \\
            & Gen $\rightarrow$ Mem & 33.8\% & 35.8\% & 30.4\% \\
        \hline
        \multirow{2}{*}{{Arithmetic addition}} 
            & Mem $\rightarrow$ Gen & 70.3\% & 28.1\% & 1.6\%  \\
            & Gen $\rightarrow$ Mem & 14.7\% & 67.6\% & 17.7\% \\
        \hline
    \end{tabular}}
    \caption{Behavioral shift observed in GPT-2 after applying inference-time intervention.}
    \label{tab:behavior_shift_GPT2}
\end{table*}

\begin{table*}[t]
    \centering
    \resizebox{0.65\textwidth}{!}{
    \begin{tabular}{|c|c|c|c|c|}
        \hline
        \textbf{Task} & \textbf{Direction} & \textbf{\% Gen} & \textbf{\% Mem} & \textbf{\% Other} \\
        \hline
        \multirow{2}{*}{{In-context inference}} 
            & Mem $\rightarrow$ Gen & 65.9\% & 19.5\%  & 14.6\% \\
            & Gen $\rightarrow$ Mem & 19.3\% & 50.9\% & 29.8\% \\
        \hline
        \multirow{2}{*}{{Arithmetic addition}} 
            & Mem $\rightarrow$ Gen & 92.3\% & 0\%  & 7.7\% \\
            & Gen $\rightarrow$ Mem & 0\% & 66.7\% & 33.3\% \\
        \hline
    \end{tabular}}
    \caption{Behavioral shift observed in LLaMA 3.2 after applying inference-time intervention.}
    \label{tab:behavior_shift_llama}
\end{table*}

\section{Generalizability of Behavior-Controlling Neurons}
\subsection{Attribution of NMD to Pretrained Base vs. LoRA Adapter}

To evaluate whether the identified memorization- and generalization-associated neurons are merely artifacts of overfitting to a specific dataset, or reflect more generalizable patterns, we begin by examining their distribution across the pretrained base model and the LoRA adapter in fine-tuned LLaMA models. Specifically, we investigate whether the observed NMD primarily originate from the frozen base model or the LoRA adapter components.

\begin{figure}[t]
    \centering
    \begin{minipage}{0.48\textwidth}
        \centering
        \includegraphics[width=1\linewidth]{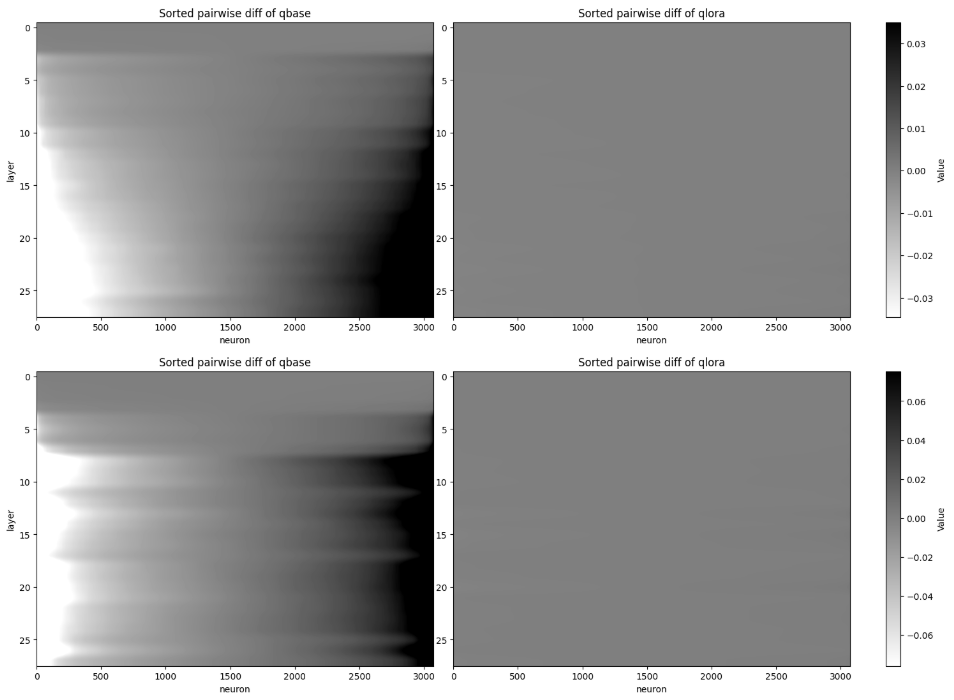}
        \caption{Neuron-wise mean differences (NMD) for the query and value projections of the base model and LoRA adapters. Left: base model; Right: LoRA adapter. Top: arithmetic addition; Bottom: in-context inference.}
        \label{fig:baselora_nmd}
    \end{minipage}
\end{figure}

In our setup, we apply LoRA adapters to the query and value projections of each transformer layer in LLaMA 3.2-3B. We compute NMD values separately for the query and value projections in both the base model and the adapter. As shown in Figure~\ref{fig:baselora_nmd}, the neurons with high NMD values are overwhelmingly concentrated in the base model, with the adapter exhibiting only minor NMD magnitudes across tasks.

This suggests that behavior-associated signals originate in the pretrained base model and persist through fine-tuning. To evaluate whether the FFN-related neurons used in our inference-time intervention exhibit consistent differentiation, we next assess their stability under different training seeds (intra-task retraining) and their transferability across tasks (inter-task generalization).

Note that While the projection-layer NMD analysis provides insight into where behavior signals originate, our consistency evaluations focus on the feed-forward layer neurons used in inference-time intervention.



\subsection{Intra-Task Consistency}
\label{sec: In-task overlap}

\begin{table*}[!t]
    \centering
    \resizebox{0.62\textwidth}{!}{
    \begin{tabular}{|c|c|c|c|c|}
        \hline
        \textbf{Task} & \textbf{Direction} & \textbf{\% Gen} & \textbf{\% Mem} & \textbf{\% Other} \\
        \hline
        \multirow{3}{*}{\shortstack{In-context inference}} 
            & Mem $\rightarrow$ Gen & 62.5\% & 35.7\%  & 1.8\% \\
            & Gen $\rightarrow$ Mem & 38.1\% & 54.8\%  & 7.1\% \\
            & Random & 0\% & 0\% & 100\% \\
        \hline
        \multirow{3}{*}{\shortstack{Arithmetic addition}} 
            & Mem $\rightarrow$ Gen & 77.1\% & 22.9\%  & 0\% \\
            & Gen $\rightarrow$ Mem & 40.7\% & 57.4\%  & 1.9\% \\
            & Random & 0\% & 0\% & 100\% \\
        \hline
    \end{tabular}}
    \caption{Intra-task behavioral shift: applying behavior-controlling neurons identified from the original LoRA adapter to a retrained adapter.}
\label{tab:retrain_ITI}
\end{table*}

\begin{table*}[!t]
    \centering
    \resizebox{0.66\textwidth}{!}{
    \begin{tabular}{|c|c|c|c|c|}
        \hline
        \textbf{Task} & \textbf{Direction} & \textbf{\% Gen} & \textbf{\% Mem} & \textbf{\% Other} \\
        \hline
        \multirow{3}{*}{\shortstack{In-context inference \\ (Arithmetic addition NMD)}} 
            & Mem $\rightarrow$ Gen & 65.9\% & 29.3\%  & 4.8\% \\
            & Gen $\rightarrow$ Mem & 63.2\% & 22.8\% & 14\% \\
            & Random & 0\% & 0\% & 100\% \\
        \hline
        \multirow{3}{*}{\shortstack{Arithmetic addition \\ (In-context inference NMD)}} 
            & Mem $\rightarrow$ Gen & 15.4\% & 69.2\%  & 15.4\% \\
            & Gen $\rightarrow$ Mem & 87\% & 9.3\% & 3.7\% \\
            & Random & 0\% & 0\% & 100\% \\
        \hline
    \end{tabular}}
    \caption{Inter-task behavioral shift: applying behavior-controlling neurons identified from the arithmetic addition task to the in-context inference task, and vice versa.}
    \label{tab:cross_task_ITI}
\end{table*}

Given that behavior-controlling neurons are primarily located in the pretrained base model, we examine whether these neurons remain effective when applied to independently retrained adapters on the same task.

To assess this, we retrain a new LoRA adapter on both the in-context inference and arithmetic addition tasks using identical data and hyperparameters, varying only the random seed. We then apply the same inference-time intervention (ITI) procedure as described in Section~\ref{sec: Inference-Time Intervention}, using the behavior-controlling neurons identified from the original adapter. 

As shown in Table~\ref{tab:retrain_ITI}, these neurons continue to steer model behavior in the retrained adapters. On the in-context inference task, success rates were 62.5\% for memorization-to-generalization and 54.8\% in the reverse direction. On the arithmetic addition task, the corresponding rates were 77.1\% and 57.4\%.

To contextualize these results, we introduce a random intervention baseline. We randomly select the same number of neurons as in the original intervention and apply weight shifts sampled uniformly from \([-v, v]\), where \(v\) is the maximum absolute shift used in the original ITI. This baseline consistently failed to steer behavior (0\% success), indicating that the observed effects are not due to arbitrary perturbation.

These results demonstrate that the identified neurons are not merely artifacts of a single training run, but generalize across adapters retrained under different initialization.

\subsection{Inter-Task Transferability}

Having established intra-task consistency, we next examine whether behavior-controlling neurons generalize across tasks. Specifically, we apply neurons identified from one task (e.g., arithmetic addition) to a different task (e.g., in-context inference), using the same inference-time intervention procedure.

Table~\ref{tab:cross_task_ITI} presents the results. Applying neurons from the arithmetic addition task to in-context inference led to a 65.9\% success rate when shifting from memorization to generalization, and 22.8\% in the reverse. Conversely, neurons identified from the in-context inference task were markedly less effective when applied to arithmetic addition, with success rates of only 15.4\% and 9.3\% respectively.

A random intervention baseline, constructed as in the intra-task experiment, again yielded no meaningful behavioral shifts, confirming that the observed effects stem from targeted neuron selection. Notably, even in the less effective transfer direction (in-context → arithmetic), the behavior shift results still outperformed the random baseline, suggesting that the selected neurons retain a weak but non-trivial steering capacity.

We hypothesize that the arithmetic addition task—being more structurally constrained—exhibits clearer neuron-level specialization for memorization and generalization, resulting in behavior-controlling neurons that generalize more effectively. In contrast, the in-context inference task may yield neurons with lower specificity, limiting their transferability. Developing improved neuron selection methods that capture higher behavioral specificity—such as combining multiple NMD metrics, leveraging attribution techniques, or performing task-aligned neuron clustering—remains an important direction for enhancing inter-task generalization.

\section{Conclusion}

This work investigates how memorization and generalization manifest within the internal structure of large language models. By identifying and manipulating behavior-associated neurons, we show that it is possible to steer model behavior at inference time and that these neurons exhibit generalizability across retraining and tasks.

Our findings uncover behavior-specific neuron-level structures that differentiate memorization and generalization within LLMs. By identifying, verifying, and manipulating these neurons, we provide empirical evidence that memorization and generalization are not just emergent capabilities, but are encoded in separable neural pathways. This opens new directions for understanding and regulating the balance between rote recall and contextual reasoning in language models. We hope this work serves as a foundation for future research focused on identifying, controlling, and  rebalancing memorization and generalization in LLMs.

\clearpage
\section*{Limitations}
While our findings shed light on the neuron-level mechanisms underlying memorization and generalization in LLMs, several limitations should be noted:

\paragraph{Model Scope.} Our experiments are limited to GPT-2 (trained from scratch) and LLaMA 3.2 with LoRA fine-tuning. While these represent both full-parameter and adapter-based training regimes, broader validation on diverse architectures and scales is necessary to assess generality.

\paragraph{Task Diversity.} We evaluate neuron behavior using only two task types—induction-style in-context inference and arithmetic addition. These are structurally distinct but do not fully capture the breadth of language tasks. Future work should examine more varied tasks, such as QA, commonsense reasoning, or dialogue.

\paragraph{Definition of Behavior.} While memorization is precisely defined via the injection of fixed input-output mappings, the operationalization of generalization is comparatively less complete. Our setup captures two specific forms of generalization, but not all forms.

\paragraph{Simplified Intervention Mechanism.} Our inference-time intervention uses a straightforward linear neuron-shifting method guided by correlation and NMD. While effective, it remains unclear whether this is the most optimal or efficient approach. More advanced strategies (e.g., ones mentioned in related works) warrant exploration.
\bibliography{main}
\clearpage
\appendix

\section*{Supplementary Materials: Training Configurations}

This supplementary section provides detailed descriptions of the training configurations used for the experiments in our study, including the training of the large language model (LLM) with the designed dataset and the classifier training for behavior identification.

\subsection*{1. Training LLM with Designed Dataset}

\paragraph{Model Architecture} We utilized GPT-2, GPT-2-medium \citep{radford2019language} and LLaMA 3.2-3B \citep{grattafiori2024llama3herdmodels} for our experiments, as described in Section~\ref{sec: Neuron Differentiation} of the main paper.

\paragraph{Dataset Design} The training datasets were specifically designed to include both memorization-specific and generalization-specific examples, as described in Section~\ref{sec: dataset design}.

\paragraph{Data Generation}

\begin{enumerate}
    \item In-Context Inference
    \begin{enumerate}
        \item Configuration Details
        \begin{itemize}
            \item Name set: 26 names
            \item Role set: 40 roles
            \item Color set: 24 colors. Each name co-occurs with 5 colors
        \end{itemize}

        \item Generation Process
        \begin{itemize}
            \item Randomly select a target name, role, and color
            \item Randomly select other names, roles, and colors to construct a coherent in-context inference story
        \end{itemize}
    \end{enumerate}

    \item Arithmetic Addition
    \begin{enumerate}
        \item Memorization Data
        \begin{itemize}
            \item Memorization pattern: 10 fixed pairs of two numbers in [1, 999]
            \item Generation process:
            \begin{itemize}
                \item Randomly select one of the memorization patterns
                \item Combine it with two randomly selected numbers in [1, 999] to form a four-number addition task
                \item Output is labeled with a special memorization token
            \end{itemize}
        \end{itemize}

        \item Generalization Data
        \begin{itemize}
            \item Generation process:
            \begin{itemize}
                \item Randomly select four numbers in [1, 999]
                \item Ensure that the 3rd and 4th numbers do not match any pair in the memorization patterns
            \end{itemize}
        \end{itemize}

        \item Sampling Probability
        \begin{itemize}
            \item In each round of training data generation:
            \begin{itemize}
                \item Memorization data is sampled with 1\% probability
                \item Generalization data is sampled with 99\% probability
            \end{itemize}
        \end{itemize}
    \end{enumerate}
\end{enumerate}

\paragraph{Training Details} The models were trained using the following configuration: \begin{itemize} 
\item \textbf{Training Algorithm:} Adam optimizer with a learning rate of $5 \times 10^{-5}$. 

\item \textbf{Batch Size:} 32 samples per batch. 

\item \textbf{LoRA Configuration:}  
For LLaMA 3.2 with LoRA fine-tuning, we set the following hyperparameters for both in-context inference and arithmetic addition tasks:
\begin{itemize}
        \item {LoRA alpha}: 32
        \item {LoRA dropout}: 0.1
        \item {Rank}: 8
        \item {Target modules}: [\texttt{"q\_proj"}, \texttt{"v\_proj"}]
\end{itemize}

\item \textbf{Model Choices:} For GPT-2 train-from-scratch scenarios, we use vanilla GPT-2 for arithmetic addition task, however, we upgraded to GPT-2 Medium for in-context inference task since vanilla GPT-2 struggles on it. For Llama with LoRA scenarios, both tasks are addressed with Llama 3.2 + LoRA fine-tuning.

\item \textbf{Training Steps:} Real-time generated training data with unlimited training steps and stop when the model demonstrates both memorization and generalization ability. Specifically, for in-context inference, we stop when LLM shows 28\% memorization and 55\% generalization output on the test data; for arithmetic addition, we stop when LLM shows 62\% memorization and 38\% generalization output on the test data.

\item  \textbf{Other:} For arithmetic addition, in order to make gpt-2 learn the task, we use the chain-of-thought approach propsed in \cite{lee2023teaching}.

\end{itemize}

\subsection*{2. Classifier Training for Behavior Prediction}

\paragraph{Classifier Input Representation} The classifier was trained to predict whether the model would engage in memorization or generalization based on the hidden states extracted from each layer of the LLM. The hidden states were extracted as described in Section~\ref{model representation extraction}.

\paragraph{Dataset Preparation} The training dataset for the classifier consisted of pairwise hidden states labeled as either "memorization" or "generalization." These hidden states were extracted from the LLM while processing the input scenarios designed to induce either behavior, as explained in Section~\ref{sec: Model Representations for Generalization and Memorization}.

\paragraph{Training Configuration} The classifiers were trained with the following configuration: \begin{itemize} \item \textbf{Classifier Architecture:} A multi-layer perceptron (MLP) with two hidden layers. For in-context inference, each layer contains twice the number of neurons as the model’s per-layer hidden state size (i.e., 2×hidden size); for arithmetic addition, each layer also contains twice the per-layer hidden state size. Both tasks use ReLU activation. \item \textbf{Training Algorithm:} Adam optimizer with a learning rate of $1 \times 10^{-5}$. \item \textbf{Batch Size:} 32 samples per batch. \item \textbf{Training Epochs:} 100 epochs with early stopping based on the validation accuracy. \item \textbf{Loss Function:} Binary cross-entropy loss. \end{itemize}

\subsection*{3. Pairwise Model Representation Dataset}
The size of each collected pairwise datasets are as follows:
\begin{itemize}
    \item \textbf{GPT2 \& in-context inference}: 80000 pairs
    \item \textbf{GPT2-medium \& arithmetic addition}: 80000 pairs
    \item \textbf{llama 3.2 \& in-context inference}: 13000 pairs
    \item \textbf{llama 3.2 \& arithmetic addition}: 6500 pairs
\end{itemize}

\subsection*{4. Hyperparameter Tuning of Inference-Time Intervention} 

The intervention involves two key hyperparameters:
\begin{itemize}
    \item \textbf{topN}: The ratio of neurons to intervene in, selected based on the highest correlation coefficients across all layers.
    \item \textbf{alpha}: The scaling factor applied to the NMD during the intervention, determining the extent of the adjustment.
\end{itemize}

If \texttt{topN} or \texttt{alpha} are too small, the intervention may not yield significant changes in the model's behavior. Conversely, if \texttt{topN} or \texttt{alpha} are too large, the intervention may excessively perturb the model, drastically altering the normal inference process. To address this, we perform a grid search to determine suitable values for \texttt{topN} and \texttt{alpha} for each task.

Given the original hidden state vector $h \in \mathbb{R}^d$ at a particular layer, we apply the intervention by modifying a subset of neurons indexed by $\mathcal{I}_{\text{topN}}$, which corresponds to the topN\% neurons ranked by absolute correlation with the target behavior. For each neuron $i \in \mathcal{I}_{\text{topN}}$, we apply a signed shift proportional to its neuron-wise mean difference (NMD) value:

\[
h_i \leftarrow h_i + \alpha \cdot \text{sign}(\rho_i) \cdot |\text{NMD}_i|
\]

Here, $\rho_i$ denotes the Pearson correlation coefficient between neuron $i$ and the target behavior (memorization or generalization), and $\alpha$ is the global scaling factor. All other neurons remain unmodified. This intervention is applied layer-wise across the model.

\end{document}